\documentclass[10pt,letterpaper]{article}
\usepackage[top=0.85in,left=2.75in,footskip=0.75in]{geometry}

\usepackage{amsmath,amssymb}

\usepackage{changepage}

\usepackage{textcomp,marvosym}

\usepackage{cite}

\usepackage{nameref,hyperref}

\usepackage[nopatch=eqnum]{microtype}
\DisableLigatures[f]{encoding = *, family = * }

\usepackage[table]{xcolor}

\usepackage{array}

\newcolumntype{+}{!{\vrule width 2pt}}

\newlength\savedwidth

\raggedright
\usepackage[aboveskip=1pt,labelfont=bf,labelsep=period,justification=raggedright,singlelinecheck=off]{caption}

\makeatletter
\renewcommand{\@biblabel}[1]{\quad#1.}
\makeatother

\usepackage{lastpage,fancyhdr,graphicx}
\usepackage{epstopdf}
\usepackage{multirow}
\usepackage{graphicx}
\usepackage{ragged2e}

\fancyheadoffset[L]{2.25in}
\fancyfootoffset[L]{2.25in}
\usepackage{fancyhdr}
\begin{document}
\justify
\vspace*{0.2in}

\begin{flushleft}
{\Large
\textbf\newline{AI-Driven Healthcare: A Review on Ensuring Fairness and Mitigating Bias} 
}
\newline
\\
Sribala Vidyadhari Chinta\textsuperscript{1},
Zichong Wang\textsuperscript{1},
Avash Palikhe\textsuperscript{1},
Xingyu Zhang\textsuperscript{2},
Ayesha Kashif\textsuperscript{3},
Monique Antoinette Smith\textsuperscript{4},
Jun Liu\textsuperscript{5*},
Wenbin Zhang\textsuperscript{1*}
\\
\bigskip
\textbf{1} Florida International University, Miami, Florida, United States of America
\\
\textbf{2} University of Pittsburgh, Pittsburgh, Pennsylvania, United States of America
\\
\textbf{3} Jose Marti MAST 6-12 Academy, Hialeah, Florida, United States of America
\\
\textbf{4} Emory University, Atlanta, Georgia, United States of America
\\
\textbf{5} Carnegie Mellon University, Pittsburgh, Pennsylvania, United States of America
\\
\bigskip

%
%





*Corresponding authors: liujun@cmu.edu, wenbin.zhang@fiu.edu

\end{flushleft}

\section*{Abstract}
Artificial intelligence (AI) is rapidly advancing in healthcare, enhancing the efficiency and effectiveness of services across various specialties, including cardiology, ophthalmology, dermatology, emergency medicine, etc. AI applications have significantly improved diagnostic accuracy, treatment personalization, and patient outcome predictions by leveraging technologies such as machine learning, neural networks, and natural language processing. However, these advancements also introduce substantial ethical and fairness challenges, particularly related to biases in data and algorithms. These biases can lead to disparities in healthcare delivery, affecting diagnostic accuracy and treatment outcomes across different demographic groups. This review paper examines the integration of AI in healthcare, highlighting critical challenges related to bias and exploring strategies for mitigation. We emphasize the necessity of diverse datasets, fairness-aware algorithms, and regulatory frameworks to ensure equitable healthcare delivery. The paper concludes with recommendations for future research, advocating for interdisciplinary approaches, transparency in AI decision-making, and the development of innovative and inclusive AI applications.

\section*{Author summary}
In this paper, we investigate the rapid advancement of artificial intelligence (AI) in healthcare, focusing on its role in improving efficiency and effectiveness across specialties such as cardiology, ophthalmology, and dermatology. We note that AI technologies enhance diagnostic accuracy, treatment personalization, and patient outcome predictions. However, these developments also pose significant ethical challenges, particularly concerning biases in data and algorithms that can create disparities in healthcare delivery. We examine the integration of AI in healthcare, highlighting the critical challenges related to bias and exploring strategies for mitigation. We emphasize the need for diverse datasets, fairness-aware algorithms, and regulatory frameworks to ensure equitable healthcare delivery. Our paper concludes with recommendations for future research, advocating for interdisciplinary approaches, transparency in AI decision-making, and the development of innovative and inclusive AI applications.


\section*{Introduction}
Artificial intelligence (AI) is revolutionizing modern healthcare, dramatically transforming the ways we diagnose, treat, and manage diseases. The integration of AI into healthcare began in the late 20th century with systems like MYCIN~\cite{shortliffe1975computer} in the 1970s, which helped diagnose infections and recommend antibiotics, and CADUCEUS~\cite{miller1984internist} in the 1980s, which emulated human diagnostic reasoning. These early systems laid the groundwork for today’s advanced machine learning and deep learning techniques, which now significantly enhance diagnostic accuracy, treatment personalization, and patient outcome predictions.

As AI technologies have advanced, their impact on healthcare has grown exponentially. Modern AI applications, particularly deep learning, have enhanced image recognition, significantly improving diagnostic accuracy in fields such as radiology and pathology~\cite{esteva2019guide}. Predictive analytics, powered by AI, are essential in patient monitoring and management, using real-time data to forecast potential patient deteriorations~\cite{rajkomar2019machine}. Additionally, natural language processing (NLP) tools have revolutionized the handling of unstructured data, improving the functionality of electronic health record systems and facilitating more comprehensive patient care~\cite{jiang2017artificial}.

Several key algorithms and technologies underpin these advancements. Neural networks, particularly convolutional neural networks (CNNs), are extensively used for medical image analysis, aiding in the detection and characterization of various pathological findings~\cite{litjens2017survey}. Decision support systems incorporate diverse data, including genetic profiles and prior health records, to optimize treatment strategies~\cite{kourou2015machine}. Among the notable AI tools, IBM Watson stands out for its application in cancer treatment, although its widespread adoption faces challenges~\cite{strickland2019ibm}.

The integration of AI into healthcare has the potential to enhance diagnostic and operational efficiencies while assisting in reducing human error~\cite{jha2016adapting}. This is due to AI's ability to manage and analyze large datasets, enabling more effective resource allocation and efficient patient scheduling, both of which can lead to improved patient outcomes and satisfaction~\cite{feldman2012BigDI}. In addition, AI-driven predictive models have proven useful in public health for tracking disease patterns and aiding in the management of epidemics, as seen during the COVID-19 pandemic~\cite{bachtiger2020machine}. These capabilities demonstrate AI's potential to enhance healthcare delivery.

 However, alongside these advancements, substantial ethical and fairness challenges have emerged. Biases embedded within training data can lead to skewed AI models, resulting in disparities in healthcare outcomes across different demographic groups~\cite{obermeyer2019dissecting}. For instance, an algorithm used in US hospitals was biased against black patients in resource allocation~\cite{chen2019can},  and dermatological AI showed lower diagnostic accuracy for conditions like melanoma in darker-skinned individuals due to training primarily on fair-skinned images ~\cite{adamson2018machine}.  Similarly, AI tools for diagnosing depression have faced challenges when applied across different linguistic and cultural backgrounds because they were primarily trained on English-speaking, Western populations, leading to potential misdiagnoses in non-Western patients~\cite{parikh2019addressing}. AI models trained predominantly on data from specific populations have exhibited lower diagnostic accuracy for underrepresented groups, exacerbating existing health disparities. These examples highlight the urgent need for diverse datasets and transparent AI systems to ensure fairness and equity in healthcare AI applications. 

Efforts to address these challenges have increasingly gained attention from governments and regulatory bodies. Recognizing the potential of AI to both improve and exacerbate healthcare disparities, policymakers have taken initial steps toward creating safeguards. The European Union's General Data Protection Regulation (GDPR), for example, provides a framework for ethical considerations in AI applications by addressing issues like data privacy and transparency~\cite{voigt2017eu}. In the United States, the Food and Drug Administration (FDA) has begun implementing guidelines to evaluate the safety and effectiveness of medical AI systems~\cite{food2019proposed}. While these initiatives mark critical progress, they primarily focus on high-level governance and oversight, leaving significant gaps in addressing the technical challenges of bias detection and mitigation. Without comprehensive strategies that integrate policy, technical methodologies, and ethical considerations, these efforts risk being insufficient to tackle the systemic biases present in AI systems.

To bridge these gaps, this paper provides a significant contribution to the field of AI-driven healthcare by offering a comprehensive framework for understanding and addressing fairness and bias. It categorizes biases across the machine learning pipeline and aligns them with targeted detection and mitigation strategies. By integrating technical, ethical, and policy perspectives, the paper addresses key challenges in creating equitable AI systems and highlights actionable solutions. Through real-world examples from diverse healthcare domains, it bridges theoretical concepts with practical implementation. This work not only deepens the understanding of bias in healthcare AI but also provides researchers, practitioners, and policymakers with valuable tools and strategies for fostering fairness and inclusivity in AI applications.

The paper is structured to help readers follow its key contributions. \textbf{Section \ref{sec:sec_1}}, discusses the applications of AI in various healthcare dimensions. The biases exhibited by those AI applications, their root causes, potential consequences, and fairness metrics and trade-offs have been discussed in \textbf{Section \ref{sec:sec_2}}. \textbf{Section \ref{sec:sec_3}} explores various approaches to address and mitigate bias. \textbf{Section \ref{sec:sec_4}} identifies research gaps and future directions for enhancing AI in healthcare. This paper is concluded in \textbf{Section \ref{sec:sec_7}}.

\section{Applications of AI in Healthcare}
\label{sec:sec_1}
AI has emerged as a transformative force in healthcare, leveraging advanced algorithms, data analytics, and machine learning techniques. By processing vast amounts of health data with remarkable speed and precision, AI systems are enhancing diagnostic accuracy, enabling personalized treatment plans, and ultimately elevating patient outcomes. This section explores the diverse applications of AI across several key areas of healthcare, highlighting its impact and potential for future innovations.
\subsection{Cardiology}
AI has the potential to transform cardiology by enhancing diagnostic accuracy, supporting personalized treatment, and contributing to improved patient care~\cite{johnson2018artificial,krittanawong2017artificial}. Specifically, AI algorithms have proven effective in analyzing cardiovascular data to aid in the early detection and diagnosis of conditions such as arrhythmias, heart failure, and coronary artery disease~\cite{attia2019screening,hannun2019cardiologist}. For instance, machine learning models have shown high precision in interpreting echocardiograms, at times  surpassing human experts in diagnostic speed and accuracy~\cite{madani2018fast}. Additionally, AI holds potential for managing cardiovascular risk by integrating diverse patient data to predict individual risk factors~\cite{ambale2015cardiac}. It also contributes to the development of personalized treatment plans by analyzing patient data and predicting responses to various treatment modalities, optimizing therapeutic decisions~\cite{krittanawong2019deep}. In oncology, AI tools like MesoNet have been used to predict patient survival by analyzing whole-slide digitized images. These tools have demonstrated greater accuracy in predicting patient survival compared to current pathology practices, offering new insights and potentially guiding better treatment decisions~\cite{courtiol2019mesothelioma}. However, the integration of AI in clinical practice requires rigorous validation and careful ethical considerations to prioritize patient safety and uphold data security~\cite{farhud2021ethical}. The continuous evolution of AI technologies holds promise for advancing diagnostic tools and therapeutic strategies, potentially improving precision and predictive capabilities in whole-person care. This could include recommending lifestyle changes tailored to individual contexts alongside medications adjusted to their genomic profiles~\cite{fuster2011global}.  Moreover, AI could play a role in supporting health equity by facilitating access to advanced cardiac care in underserved areas through telemedicine platforms and remote monitoring, which could help reduce disparities in cardiovascular health outcomes~\cite{topol2019high}.
\subsection{Ophthalmology}
AI has demonstrated significant potential in ophthalmology by improving diagnostic accuracy for common eye diseases and streamlining clinical workflows. Deep learning models have been particularly effective in analyzing complex visual data from imaging techniques such as fundus photography and optical coherence tomography (OCT) \cite{ting2017development}. These models are very good at diagnosing common eye diseases like diabetic retinopathy, glaucoma, and age-related macular degeneration, sometimes achieving expert-level sensitivity and specificity~\cite{gulshan2016development}. Automated analysis using AI not only speeds up the diagnostic process but also reduces human error, facilitating earlier and more precise interventions~\cite{de2018clinically}. Furthermore, AI's predictive capabilities are being harnessed to forecast disease progression, which is crucial for conditions like glaucoma, where early detection can prevent severe vision loss~\cite{litjens2017survey}. Using AI tools in clinical practice also includes using AI-driven decision support systems that help plan treatments by guessing how different types of treatments will work~\cite{balyen2019promising}. Despite these advancements, challenges remain, including data privacy concerns, the need for large annotated datasets for training algorithms, and the integration of AI into existing clinical workflows~\cite{kermany2018identifying}. However, ongoing research and collaboration between AI technologists and ophthalmic experts are likely to overcome these obstacles, solidifying AI’s role in modern ophthalmology~\cite{abramoff2018pivotal}. This further opens the frontier for the expansion of access to ophthalmological care in existing deserts through technology-enabled care models and non-specialist operators at the point of service, thereby addressing disparities in eye care availability and quality.

\subsection{Dermatology}
AI is advancing dermatology by improving diagnostic accuracy, personalizing treatments, and streamlining patient management. Deep learning models, such as CNNs, are effective in diagnosing skin cancer, demonstrating performance comparable to dermatologists in identifying melanomas and other skin conditions \cite{esteva2017dermatologist,han2018classification}. Moreover, AI applications in dermatology extend beyond cancer detection to managing chronic conditions like psoriasis and atopic dermatitis. These algorithms assist in monitoring disease progression and treatment response, enhancing decision-making in telemedicine and clinical workflows \cite{tschandl2020human}. AI systems are also utilized in cosmetic dermatology, optimizing treatment recommendations based on individual facial analysis~\cite{gomolin2020artificial}. Ensuring the efficacy and safety of AI tools requires rigorous validation processes with diverse datasets to mitigate bias and enhance generalizability~\cite{adamson2018machine}. Integration into clinical workflows and user training is crucial to maximizing the benefits of AI in dermatological practice~\cite{liu2020deep}. As AI continues to evolve, continuous collaboration between technologists and clinicians is essential to address ethical considerations and improve patient outcomes in dermatology. Importantly, AI can democratize access to high-quality dermatological care by enabling remote consultations and diagnostics, thus reducing barriers for individuals in remote or underserved regions.

\subsection{Neurology}
AI has contributed to advancements in neurology by improving diagnostic accuracy and supporting research efforts. Techniques like machine learning and deep learning assist in analyzing imaging data for disorders such as Alzheimer’s and epilepsy \cite{lecun2015deep,jiang2017artificial}. These technologies enhance the interpretation of MRI and CT scans, often achieving higher accuracy than traditional methods~\cite{litjens2017survey}. In treatment, AI algorithms personalize therapies based on patient data, optimizing outcomes for diseases like Parkinson’s~\cite{dorsey2018parkinson}. Additionally, AI has advanced neuroprosthetics by enabling adaptive brain-machine interfaces (BMIs), which, when combined with physiotherapy, show potential to improve motor function in patients with severe motor disabilities \cite{ramos2013brain}. Predictive analytics in healthcare highlight the importance of transparent algorithms for reliable validation and monitoring, which support applications in neurology, such as tracking disease progression in conditions like multiple sclerosis \cite{van2019predictive}. AI's applications extend to research, facilitating the understanding of complex neurological phenomena and the development of innovative treatments~\cite{bzdok2018machine}. For instance, StrokeSight, an EEG-based system, demonstrates how AI can improve neurology by providing fast, affordable stroke diagnoses and personalized treatments using deep learning~\cite{kalahasty2022strokesight}. Overall, AI's integration into neurology not only enhances clinical practices but also opens new avenues to address rare diseases and develop more finely tuned medications tailored to individual patients~\cite{fornito2016fundamentals}.

\subsection{Radiology and Cancer Treatments}
AI is advancing radiology and cancer treatment by enhancing imaging accuracy, enabling personalized therapies, and improving diagnostic workflows. Deep learning algorithms have shown significant progress in interpreting medical images, supporting the early detection and diagnosis of cancers such as breast and lung cancer \cite{hosny2018artificial}. Machine learning models assist radiologists by identifying complex patterns in imaging data, aiding in faster and more efficient diagnoses \cite{wang2012machine}. In cancer treatment, AI assists in formulating personalized treatment plans based on patient data and predictive analytics, which can predict treatment outcomes and suggest optimal therapies~\cite{yu2018artificial}.  AI applications in radiology extend to prognostic evaluations, predicting disease progression and survival rates, refining treatment protocols and developing follow-up strategies ~\cite{bi2019artificial}. The development of AI systems capable of outperforming human experts in tasks like breast cancer prediction highlights their potential in clinical settings. These advancements are supported by large datasets and strict validation processes, ensuring reliability and clinical applicability \cite{mckinney2020international}.

\subsection{Emergency Medicine and Critical Care}

AI stands to significantly transform emergency medicine and critical care by enhancing patient triage, treatment efficacy, and time to diagnosis. As AI-powered tools like diagnostic algorithms have been incredibly good at the early identification of conditions like sepsis and its progression, there is the potential to improve patient outcomes~\cite{wardi2021predicting, brann2024sepsis}. The prediction of patient outcomes has also positioned AI to facilitate the optimization of critical resource allocation. During the COVID-19 pandemic, AI-based self-triage tools had a critical role to play in predicting cases and hospitalizations at the population level~\cite{lee2022use}. AI-powered chatbots and virtual assistants are streamlining patient intake and symptom assessment~\cite{miner2020chatbots, semigran2015evaluation}. Furthermore, AI’s role in triaging patients based on urgency has revolutionized care prioritization, improving ED operations and patient flow~\cite{dong2020interactive, sterling2020prediction}. These advancements underscore AI’s potential to not only enhance emergency medical services but also pave the way for a more efficient, accurate, patient-centered healthcare system that extends care beyond traditional brick-and-mortar facilities.

\section{Fairness Concerns in Healthcare}
\label{sec:sec_2}
This section explores the fairness concerns associated with the use of AI in healthcare through three key areas. It begins by examining the sources of bias that can emerge at various stages of the machine learning pipeline. Next, it discusses the potential consequences of these biases, including inequitable healthcare outcomes, loss of trust, and ethical and legal challenges. Finally, it introduces fairness metrics and trade-offs, which provide a framework for addressing bias while balancing the inherent tensions between fairness and performance in healthcare applications. By delving into these topics, this section aims to underscore the critical need to address biases to ensure AI technologies contribute to fair and equitable healthcare for all.

\subsection{Sources of Bias}
\label{sec:sec_sources}

Bias in AI healthcare systems can arise at multiple stages of the machine learning (ML) pipeline, necessitating phase-specific mitigation strategies. To address this comprehensively, researchers have developed three primary approaches, each targeting a distinct phase of the pipeline: (i) pre-processing, which focuses on reducing bias in the input data before model training~\cite{caton2024fairness}; (ii) in-processing, which applies fairness constraints during the training phase~\cite{hort2024biasmitigation}; and (iii) post-processing, which modifies model outputs to enhance fairness~\cite{feldman2021endtoend}. We adopt this three-stage framework to capture the end-to-end process of data collection, model development, and decision-making in AI healthcare systems. This structure, widely used in the fairness literature, supports systematic bias analysis and enables targeted detection and mitigation strategies at each stage~\cite{mehrabi2021survey,caton2024fairness,hort2024biasmitigation,feldman2021endtoend,pessach2022fairness, pagano2023bias}. The division also facilitates clearer reasoning about when and how fairness interventions can be most effectively applied throughout the ML pipeline.

A critical first step in designing effective interventions is to identify the underlying sources of bias at each stage of the ML pipeline. This section outlines those sources, with a summary provided in Table~\ref{table:summary_table_sources_of_bias} and detailed discussions presented in the following sections.

\begin{table}[]
\centering
\caption{Summary of biases across each stage of the machine learning (ML) pipeline}
\vspace{0.5cm}
\label{table:summary_table_sources_of_bias}
\begin{tabular}{|l|l|l|p{4.8cm}|}
\hline
\textbf{Stage} & \textbf{Bias Type} & \textbf{References} & \textbf{Description} \\
\hline
\multirow{5}{*}{Pre-processing} 
  & Selection Bias&\mbox{\cite{ferrara2024fairness,mittermaier2023bias}}&Arises from how the data is chosen or sampled for training, leading to unrepresentative datasets.\\
\cline{2-4}
  & Measurement Bias&\mbox{\cite{mehrabi2021survey,ferrara2024fairness,gottlieb2022oxygen}}&Results from the ways features are chosen, utilized, or measured, leading to the systematic over- or under-representation of certain groups or variables in the data.\\
\cline{2-4}
  &Representation Bias&\mbox{\cite{ferrara2024fairness,adamson2018machine}}&Occurs when datasets fail to adequately capture the diversity of the populations they aim to serve. \\
\cline{2-4}
  &Explicit Bias&\mbox{\cite{mehrabi2021survey}}&Involves intentional or overt decisions that result in discriminatory patterns in data preparation or selection. \\
\cline{2-4}
  &Implicit Bias&\mbox{\cite{friedman1996bias}}&Stems from unintentional or unconscious decisions that lead to skewed datasets, frequently influenced by systemic inequities. \\
\hline
\multirow{3}{*}{In-processing} 
  & Algorithmic Bias     & \mbox{\cite{ueda2024fairness,vyas2020hidden}}          & Arises when algorithms magnify existing biases in the training data or when they are inherently biased due to their design. \\
\cline{2-4}
  & Explicit Bias        & \mbox{\cite{mehrabi2021survey}}                & Involves deliberate manipulations within the model’s algorithms or training process that introduce or amplify disparities. \\
\cline{2-4}
  & Implicit Bias        & \mbox{\cite{suresh2019framework}}                & Arises from model design decisions that unintentionally embed existing societal inequities, often due to insufficient awareness or oversight. \\
\hline
\multirow{3}{*}{Post-processing} 
  & Evaluation Bias      & \mbox{\cite{cross2024bias}}                & Arises when performance assessments fail to account for disparities across demographic groups, leading to inequities in clinical outcomes. \\
\cline{2-4}
  & Explicit Bias        & \mbox{\cite{mehrabi2021survey}}                & Involves intentional manipulation of outputs or decision thresholds that disproportionately affect specific groups. \\
\cline{2-4}
  & Implicit Bias        & \mbox{\cite{friedman1996bias}}                & Occurs when decision-making criteria unintentionally create disparities, often because important fairness factors are overlooked. \\
\hline
\end{tabular}
\label{tab:bias_summary}
\end{table}

\subsubsection{Pre-Processing Stage}

In the pre-processing stage, bias can originate from data collection, preparation, or representation before modeling begins. Because both the performance and fairness of an AI system depend on the quality of its data, bias at this stage can have widespread consequences. One common form of bias is \textit{selection bias}, which arises from how the data is chosen or sampled for training, leading to unrepresentative datasets \cite{ferrara2024fairness}. For instance, selection bias occurs when an AI model is trained predominantly on data from one demographic (e.g., white patients), while underrepresented groups, such as Black patients, have limited representation or inaccessible health records. This can lead to false negatives, fewer follow-up scans, and undiagnosed conditions, ultimately worsening health inequities for disadvantaged populations \cite{mittermaier2023bias}.

Another form of bias is \textit{measurement bias}, which results from the ways features are chosen, utilized, or measured, leading to the systematic over- or under-representation of certain groups or variables in the data \cite{mehrabi2021survey,ferrara2024fairness}. An example is the use of pulse oximeters, devices that estimate blood oxygen levels by emitting light through the skin. Research indicates that pulse oximeters often provide less accurate readings for individuals with darker skin tones, frequently overestimating their oxygen saturation levels~\cite{gottlieb2022oxygen}. This discrepancy can result in the underdiagnosis or delayed treatment of hypoxemia in these populations, rendering the models unreliable or biased in their predictions. Similarly, \textit{representation bias} occurs when datasets fail to adequately capture the diversity of the populations they aim to serve \cite{ferrara2024fairness}. For example, dermatological AI systems trained primarily on images of lighter skin tones often underperform when diagnosing conditions on darker skin \cite{adamson2018machine}. This lack of representation can lead to inequitable healthcare outcomes and limit the generalizability of the AI system.

Within pre-processing, biases can also appear explicitly or implicitly. \textit{Explicit bias} involves intentional or overt decisions that result in discriminatory patterns in data preparation or selection \cite{mehrabi2021survey}. For instance, consider a dataset for training an AI model to predict heart disease. If the dataset is curated to include only data from male patients and excludes female patients entirely, this introduces explicit bias. This exclusion is deliberate and creates a model that cannot generalize to female patients, potentially leading to disparities in healthcare outcomes. \textit{Implicit bias}, on the other hand, often stems from unintentional or unconscious decisions that lead to skewed datasets, frequently influenced by systemic inequities \cite{friedman1996bias}. For example, using historical healthcare data that disproportionately underrepresented women in clinical trials introduces an implicit bias into the AI model. The unconscious reliance on historically biased datasets perpetuates existing inequalities, such as the underdiagnosis of heart attacks in women.

\subsubsection{In-Processing Stage}

In the in-processing stage, biases can emerge during model training or can be introduced through algorithmic choices, significantly impacting an AI system’s outputs. Although the data itself can carry biases, the algorithms that learn from this data can also amplify or introduce a form of bias often referred to as \textit{algorithmic bias}. It arises when algorithms magnify existing biases in the training data or when they are inherently biased due to their design \cite{ueda2024fairness}. For example, the Vaginal Birth After Cesarean (VBAC) calculator included race-based correction factors that systematically assigned lower success probabilities to African American and Hispanic women, discouraging VBAC attempts for these groups without robust scientific justification \cite{vyas2020hidden}. Such a design flaw not only reflected existing disparities in maternal healthcare but also exacerbated them, highlighting how biased algorithms can influence decision-making in critical areas.

Explicit and implicit biases can also arise during in-processing. \textit{Explicit bias} involves deliberate manipulations within the model’s algorithms or training process that introduce or amplify disparities \cite{mehrabi2021survey}. For instance, intentionally adjusting the algorithm to prioritize predictions for one demographic group over another (e.g., optimizing cancer detection accuracy only for men) constitutes explicit bias. Such direct alterations can skew the model’s performance metrics, favoring specific groups while disadvantaging others, resulting in ethical and clinical concerns. \textit{Implicit bias}, however, arises from model design decisions that unintentionally embed existing societal inequities, often due to insufficient awareness or oversight \cite{suresh2019framework}. For example, training a diabetes prediction model using an optimization metric that inadvertently prioritizes accuracy over fairness for underserved populations can perpetuate health disparities. These unconscious design choices may cause the model to fail to serve vulnerable groups adequately.

\subsubsection{Post-Processing Stage}

Finally, the post-processing stage involves how model outputs are evaluated, interpreted, and deployed in real clinical settings. \textit{Evaluation bias} arises when performance assessments fail to account for disparities across demographic groups, leading to inequities in clinical outcomes \cite{cross2024bias}. For example, if an AI model for detecting diabetic retinopathy is evaluated primarily on data from urban hospitals, its performance metrics may overstate its accuracy and fail to reflect reduced performance in rural or underserved populations. This oversight in the evaluation process can lead to biased deployment decisions, disproportionately affecting access to reliable diagnostics in marginalized communities.

As in earlier stages, biases at this point can be explicit or implicit. \textit{Explicit bias} involves intentional manipulation of outputs or decision thresholds that disproportionately affect specific groups \cite{mehrabi2021survey}. For instance, adjusting the risk score thresholds for approving diagnostic tests based on race or socioeconomic status. Such actions, when deliberate, can reinforce discriminatory practices in clinical decision-making and resource allocation. \textit{Implicit bias}, on the other hand, occurs when decision-making criteria unintentionally create disparities, often because important fairness factors are overlooked \cite{friedman1996bias}. Consider an AI system that predicts which patients are at risk of readmission after hospital discharge. During the post-processing stage, the hospital decides to prioritize follow-up care for patients with private insurance, based on an assumption that they are more likely to comply with medical recommendations. However, this approach unintentionally introduces implicit bias, as patients with public insurance or no insurance who may face greater barriers to accessing care are deprioritized. The bias does not stem from a deliberate decision to disadvantage these groups, but rather from an unconscious oversight of fairness considerations in the post-processing criteria.

\bigskip

\noindent By examining biases according to their emergence in the ML pipeline, this structured framework facilitates clearer understanding of their origins and supports targeted mitigation strategies. Such an approach underscores the importance of addressing biases early and continuously from data collection and algorithm design to final model deployment and evaluation to ensure equitable and responsible AI-driven healthcare solutions.

\subsection{Potential consequences of biases}
Biases in AI systems, particularly in healthcare, can have profound and far-reaching consequences. These biases usually stem from the data on which the AI systems are trained, the design of the algorithms themselves, and the contexts in which they are deployed (as discussed in Section \ref{sec:sec_sources}). Here are several potential consequences of biases in AI systems in healthcare:

\textbf {Misdiagnosis and Inequitable Healthcare Outcomes:} AI integration in healthcare brings significant advancements but also risks of misdiagnosis and inequitable outcomes due to biased algorithms. AI systems often rely on non-representative data, leading to biased decision-making. For instance, Obermeyer highlighted that an algorithm used in healthcare disproportionately favored white patients over black patients because it used healthcare costs as a proxy for healthcare needs, indirectly embedding racial biases in its predictions~\cite{obermeyer2019dissecting}.
Moreover, these biases in AI can exacerbate existing healthcare disparities. Rajkomar and others discussed how AI applications, if not carefully designed and monitored, could inherit and amplify socioeconomic and racial disparities. This is particularly problematic in diagnostics, where AI systems are trained predominantly on data from specific demographic groups, potentially leading to poorer diagnostic accuracy for underrepresented groups~\cite{rajkomar2019machine}. This was evident in a study by Adamson and Smith, which found that dermatology AI systems demonstrated lower accuracy rates in skin lesion diagnosis for dark-skinned individuals compared to those with lighter skin~\cite{adamson2018machine}.
Such biased AI not only risks misdiagnosis but also contributes to inequitable healthcare outcomes by potentially steering healthcare resources away from those who may need them most. Vayena and his team emphasized the ethical imperative to ensure that AI tools in healthcare are developed with consideration for fairness and equity, advocating for diverse and inclusive data sets and algorithmic transparency to mitigate these biases~\cite{vayena2018machine}.

\textbf {Loss of trust in Healthcare Systems:} Recent studies have echoed the concern that the deployment of biased AI in healthcare could significantly undermine public trust in medical systems. Trust is foundational to the patient-provider relationship and is crucial for the effective delivery of healthcare services~\cite{mittelstadt2016ethics}. When AI tools exhibit bias, whether in diagnosis, treatment recommendations, or patient management, they can lead to misdirected care, fostering distrust among patients, particularly in marginalized communities~\cite{rajkomar2019machine}. Kherbache and his team pointed out that when patients perceive AI-driven processes as opaque or unfair, their trust in the overall healthcare system may decline~\cite{kherbache2022moral}. Incidents of AI failures that receive public attention can exacerbate this erosion of trust, leading patients to question the reliability and ethics of using AI in medical decision-making. Researchers like Kerasidou argue that trust is not only about the accuracy of AI but also its alignment with ethical principles that govern healthcare, such as beneficence and non-maleficence~\cite{kerasidou2017trust}. Furthermore, the lack of transparency in how AI models make decisions can be a major barrier to trust. Blease's study~\cite{blease2019artificial} suggests that without clear communication about how AI tools contribute to healthcare decisions, patients may become skeptical of diagnoses and treatments, fearing that their personal healthcare data could be misused or misunderstood. This potential trust deficit could have severe implications, not just for individual health outcomes but also for public health at large, as mistrust in healthcare systems can lead to lower rates of healthcare utilization, vaccine hesitancy, and poor adherence to medical advice~\cite{larson2014understanding}. Veinot highlights the need for healthcare systems to maintain high standards of accountability and transparency as they integrate AI technologies to mitigate such risks~\cite{veinot2018good}.

\textbf {Legal and Ethical Implications:} The use of biased AI in healthcare not only poses clinical risks but also entails significant legal and ethical implications. Deploying biased AI systems could lead to legal breaches of anti-discrimination laws. In the United States, for example, the Civil Rights Act and the Americans with Disabilities Act set legal standards that could be violated by biased AI algorithms, which fail to provide equitable care across different patient demographics~\cite{carter2020ethical}. Ethically, biased AI conflicts with the fundamental medical ethics principles of justice and non-maleficence, demanding fairness and avoidance of harm, respectively~\cite{char2018implementing}. Moreover, biased algorithms in healthcare could potentially expose medical practices and institutions to litigation related to malpractice or negligence, especially if these algorithms contribute to substandard care outcomes~\cite{price2019privacy}. For instance, if an AI system were to consistently provide inferior diagnostic support for certain racial groups, this could be viewed as a form of systemic negligence or malpractice~\cite{vayena2018machine}. Additionally, the ethical implications extend to the breach of patient trust and the compromise of patient autonomy. The ethical medical practice relies heavily on the principles of informed consent and respect for patient’s autonomy—principles that are challenged by opaque AI systems that do not make their decision-making processes or inherent biases clear to patients or practitioners~\cite{blease2019artificial}. These potential legal and ethical failures underscore the necessity for rigorous oversight and transparent development processes for AI in healthcare, aiming to ensure that these technologies adhere to both existing legal frameworks and ethical standards of practice.

\textbf{Resource Misallocation:} The deployment of biased AI in healthcare settings can lead to resource misallocation, a critical issue that impacts both the efficiency and fairness of medical services. Biased algorithms may misdirect resources by prioritizing certain groups over others based on flawed data inputs or biased training procedures. For example, a study by Obermeyer demonstrated how an algorithm used for managing healthcare resources inadvertently favored healthier white patients over sicker black patients due to biased data inputs that did not accurately reflect patient needs~\cite{obermeyer2019dissecting}. This misallocation can exacerbate existing healthcare disparities by diverting necessary medical attention and resources away from those who are most in need. As Chen pointed out, such disparities are not just a matter of clinical outcomes but are deeply tied to social and economic inequalities that AI tools can inadvertently perpetuate~\cite{chen2019can}. Furthermore, biased AI can influence the allocation of resources within healthcare facilities, potentially resulting in inefficiencies that strain healthcare systems. This includes misallocating medical staff, diagnostic tools, and hospital beds, which can degrade the quality of care delivered while increasing wait times and healthcare costs~\cite{rajkomar2019machine}. Resource misallocation also raises ethical questions about fairness and equity in healthcare provisioning. It challenges the ethical principle of justice, which demands that healthcare resources be distributed based on need rather than biased algorithms~\cite{char2018implementing}. This ethical breach can lead to further mistrust and reluctance among underserved populations to engage with healthcare systems, perpetuating a cycle of disadvantage.

\textbf {Stifling Innovation:} The presence of bias in AI systems used in healthcare not only affects the accuracy and fairness of medical services but can also stifle innovation. When AI models are made using biased data, they might not accurately reflect the different needs of the population. This could make these innovations less useful and applicable to a wider range of demographic groups. Liang talks about how biased training datasets in AI development make it harder for these systems to work with different types of data. This could stymie innovation by preventing it from solving larger or more complex health problems that affect many people.
Moreover, reliance on biased AI can deter investment in developing new technologies that are inclusive and equitable. Potential investors may be cautious about funding projects that might not meet regulatory standards for fairness or could lead to public backlash — a concern highlighted by Corbett-Davies and Goel, who note the legal and social implications of deploying biased AI~\cite{corbett2023measure}. In addition, the perpetuation of bias in AI could solidify existing disparities in healthcare innovation. As Benjamin notes, if innovation is directed predominantly at solving problems specific to well-represented groups, less common but equally pressing issues in underrepresented groups are neglected, thereby limiting the scope and impact of technological advancements in healthcare~\cite{benjamin2019race}. The studies of Rajkomar~\cite{rajkomar2019machine}, Hardt~\cite{hardt2024love}, and Howell~\cite{howell2024three} made things even more complicated by pointing out that biased algorithms can make it harder to find new treatments that work for everyone. This feeds a cycle of innovation that helps groups that are already overrepresented in data sets more than others. This restricted focus on innovation not only affects the equity of healthcare delivery but also limits the development of truly innovative, comprehensive healthcare solutions. These factors combined suggest that biased AI not only hinders the progression of medical technologies but also potentially locks the healthcare sector into a cycle of uneven innovation where only the needs of the majority are systematically addressed.

\subsection{Fairness Metrics and Trade-offs}

\noindent Biases in AI healthcare systems, as discussed in the previous sections, can significantly impact both the quality and equity of healthcare delivery. To address these challenges, fairness metrics provide a structured framework for evaluating and mitigating disparities in AI model predictions. Metrics such as Statistical Parity~\cite{dwork2012fairness}, Equal Opportunity~\cite{hardt2016equality}, and Equalized Odds~\cite{le2022survey} serve as foundational tools for detecting bias. However, these metrics often reflect distinct fairness priorities, leading to inherent trade-offs when applied simultaneously~\cite{saravanakumar2021impossibility}. For example, optimizing Statistical Parity to ensure equal outcomes across groups may conflict with Equal Opportunity, which focuses on ensuring equal access to positive outcomes for individuals who qualify. These conflicts underscore the complexity of achieving fairness in healthcare AI.

Furthermore, tensions arise not only between fairness metrics but also between different dimensions of fairness, such as group fairness and individual fairness. While group fairness seeks equitable treatment across demographic groups, individual fairness emphasizes similar outcomes for individuals with comparable features. These trade-offs become even more challenging in healthcare scenarios where fairness must often be balanced with performance, such as ensuring accurate predictions while maintaining equity. By navigating these conflicts, we aim to design AI systems that uphold fairness as a core principle in healthcare. To provide a clear understanding, we introduce these three key fairness metrics along with their mathematical foundations.

\bigskip

\noindent \textbf{Statistical Parity (SP)}: Statistical Parity~\cite{dwork2012fairness} ensures that the probability of a positive predicted outcome is equal across different groups defined by a sensitive attribute. Mathematically, it is defined as:

\bigskip

\begin{equation}
    SP = P(\hat{Y} = 1 \mid S = s_1) - P(\hat{Y} = 1 \mid S = s_2)
\end{equation}

\noindent where \(\hat{Y}\) is the predicted outcome and \(S\) is the sensitive attribute. Statistical Parity is satisfied when \textit{SP}= 0, indicating no disparity between the groups.

\bigskip

\noindent \textbf{Equal Opportunity (EO)}: Equal Opportunity~\cite{hardt2016equality} ensures that individuals who are likely to achieve the advantaged outcome (\textit{e.g.}, repaying a loan, being admitted to college) have an equal chance of receiving the positive prediction, regardless of their sensitive attribute (\textit{e.g.}, race, gender). Mathematically, it is defined as:

\begin{equation}
EO = P(\hat{Y} = 1 \mid S = s_1, Y = 1) - P(\hat{Y} = 1 \mid S = s_2, Y = 1)
\end{equation}

\noindent where \(\hat{Y}\) is the binary prediction, \(S\) is the sensitive attribute, and \(Y = 1\) represents the advantaged outcome. Equal Opportunity is satisfied when \(EO = 0\) indicates no disparity in predictions for the advantaged outcome across sensitive groups.

\bigskip

\noindent \textbf{Equalized Odds (Eq.Odds)}: Equalized Odds~\cite{le2022survey} ensures that for each ground truth label (Y = y), the predicted outcome \(\hat{Y}\) is equally likely across groups defined by the sensitive attribute (S). Mathematically:

\begin{equation}
\text{Eq.Odds} = \sum_{y \in \{0, 1\}} \big| P(\hat{Y} = 1 \mid S = s_1, Y = y) - P(\hat{Y} = 1 \mid S = s_2, Y = y) \big|
\end{equation}

\noindent where $\hat{Y}$ is the predicted outcome, \(S\) is the sensitive attribute, and $Y$ is the ground truth label. Equalized Odds is satisfied when \(Eq.Odds=0\), indicating no disparity across groups for all ground truth labels.



\bigskip

These group fairness metrics Statistical Parity, Equal Opportunity, and Equalized Odds reflect fundamentally different fairness priorities, leading to inherent trade-offs when attempting to satisfy them simultaneously. Statistical Parity requires predictions to be independent of sensitive attributes, ensuring equal positive outcome rates across groups, regardless of differences in their underlying characteristics, which can sometimes lead to disparities in how individuals within a group are treated. Consider a healthcare scenario where males have a higher prevalence of heart disease compared to females, enforcing Statistical Parity would mean predicting the same proportion of males and females as being at risk. However, this might conflict with Equal Opportunity, which requires equal true positive rates across groups. Achieving Statistical Parity in this scenario may lower the true positive rate for females, as increasing their positive predictions could include many individuals without heart disease. This highlights how prioritizing Statistical Parity can undermine Equal Opportunity by creating disparities in prediction accuracy within groups.

To achieve Equalized Odds, a model must ensure that true positive and false positive rates are the same across groups, which often requires adjusting prediction thresholds differently for each group. For example, if females are less likely to have heart disease compared to males, the model might lower the prediction threshold for females to classify more of them as ``at risk" increasing the true positive rate for females. At the same time, it might raise the threshold for males to reduce false positives. While these adjustments equalize predictive performance across groups, they disrupt Statistical Parity by creating unequal proportions of positive predictions for males and females, reflecting the differences in the underlying likelihood of heart disease between the groups. This trade-off shows how achieving Equalized Odds can violate Statistical Parity by altering the balance of positive predictions across groups.

The trade-offs between these fairness metrics are formalized in the Impossibility Theorem ~\cite{saravanakumar2021impossibility}, which demonstrates that it is often impossible to satisfy all fairness criteria simultaneously. For instance, achieving Statistical Parity might require selecting more from a group with a lower base rate, sacrificing Equalized Odds or Equal Opportunity, while prioritizing Equal Opportunity or Equalized Odds might allow selection rates to differ across groups, violating Statistical Parity. These inherent conflicts require careful consideration when deciding which fairness metric to apply. Ultimately, the choice of fairness metric should depend on the specific context and the ethical or practical priorities of the application. For instance, in critical healthcare scenarios like diagnosing heart disease, Equal Opportunity or Equalized Odds might be prioritized to ensure equal access to accurate treatment and minimize disparities in predictive performance. On the other hand, Statistical Parity could be more relevant in preventive care programs, where equitable outreach and representation are paramount goals~\cite{kleinberg2016tradeoffs}.

\begin{figure}[!htb]
    \centering
    \includegraphics[width=1.07\textwidth]{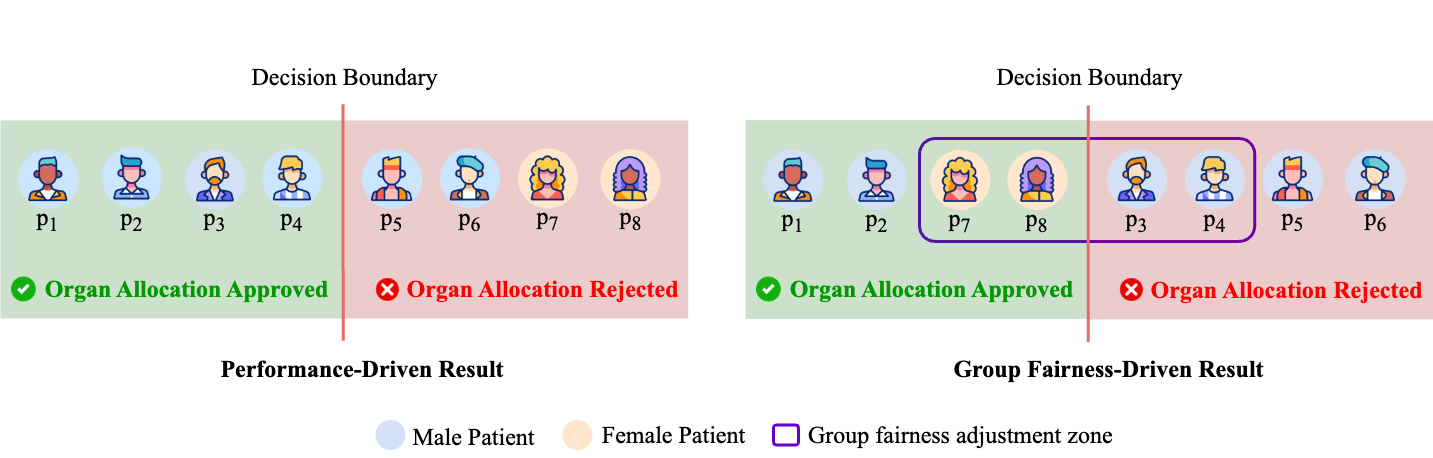}
    \vspace{0.2cm}
    \caption{Illustration of the trade-off between performance-driven result and group fairness-driven result in the context of organ allocation, where gender is the sensitive attribute.}
    \label{fig:trade-off between performance-driven result and group fairness-driven result}
\end{figure}

The conflict between fairness metrics can also be seen in the difference between individual and group fairness, which focuses on fairness at different levels. Group fairness~\cite{zhang2023fair} aims to ensure equitable treatment across different subgroups defined by sensitive attributes, \textit{i.e.}, group fairness aims to ensure that the outputs of ML algorithms are independent of the sensitive attribute. In contrast, individual fairness~\cite{wang2024individual} enforces fairness at a finer granularity by ensuring that similar individuals receive similar predictions based on their input features. Unfortunately, the simultaneous achievement of both maxima is practically impossible due to the trade-offs involved. Fig~\ref{fig:trade-off between performance-driven result and group fairness-driven result} illustrates this trade-off in the context of organ allocation, where gender is the sensitive attribute, and medical urgency determines priority. In the performance-driven result (left side of the Fig~\ref{fig:trade-off between performance-driven result and group fairness-driven result}), organs are allocated solely based on urgency and likelihood of survival, prioritizing patients $P_1$, $P_2$, $P_3$, and $P_4$ due to their higher need and better survival prospects, while rejecting $P_5$, $P_6$, $P_7$, and $P_8$ because of their lower urgency or expected benefit.  This approach maximizes utility by allocating resources to those expected to achieve the greatest overall benefit, such as improved survival rates and long-term health outcomes. However, it may inadvertently perpetuate systemic disparities between subgroups. The group fairness-driven result (right side of the Fig~\ref{fig:trade-off between performance-driven result and group fairness-driven result}) adjusts the decision boundary to ensure equal approval rates for males and females, resulting in females $P_7$ and $P_8$ being approved despite their lower urgency, while males $P_3$ and $P_4$, with higher urgency, are denied. While this adjustment addresses equity at the group level, it compromises individual fairness, as highlighted by the purple rectangle, where individuals near the decision boundary disproportionately bear the group fair loss of the entire subgroup.

Building on the discussion of trade-offs between individual and group fairness, the tension between fairness and performance further illustrates these challenges. In contexts such as organ allocation, performance-driven approaches prioritize outcomes for individuals with the highest urgency or expected benefit. However, enforcing group fairness constraints such as ensuring equal approval rates across subgroups defined by sensitive attributes like gender can lead to a reduction in overall performance. For example, in the organ allocation scenario, ensuring that approval rates are equal for males and females might result in allocating organs to individuals like $P_7$ and $P_8$ (females, lower urgency) while denying organs to $P_3$ and $P_4$ (males, higher urgency). This adjustment ensures equity at the group level but comes at the cost of reduced overall utility, as fewer lives may be saved or long-term health outcomes may be compromised. This trade-off highlights the inherent difficulty of simultaneously achieving fairness and optimal performance, as prioritizing one objective often necessitates sacrifices in the other.

\section{Addressing and Mitigating Unfairness in AI}
\label{sec:sec_3}
As AI technologies increasingly influence healthcare delivery, it becomes essential to scrutinize and refine these systems to prevent disparities in care and outcomes. This section explores various bias-detecting and mitigating strategies in AI systems in healthcare.

\subsection{Bias Detection Methods}

Bias detection is a crucial step in identifying and addressing inequities in AI systems, especially in healthcare given its high-stakes nature. This section discusses various tools and techniques for detecting bias, categorized by their roles in the pre-processing, in-processing, and post-processing stages of the machine learning (ML) pipeline. The aim is to provide a clear and actionable understanding of bias detection.

\subsubsection{Pre-Processing Stage}

The pre-processing stage focuses on detecting biases in datasets before model training begins. Identifying these biases at this stage is essential to prevent their propagation and amplification through the ML pipeline if left unchecked. Specifically, bias can be detected using \textit{Fairness Metrics Techniques}, such as Equalized Odds and Equal Opportunity Difference, which measure disparities in clinical outcomes or feature distributions across demographic groups. These metrics help identify biases in healthcare datasets, such as selection bias and measurement bias by revealing whether certain patient populaitons are disadvantaged or overrepresented. For instance, if one group consistently experiences worse outcomes in the data, these tools can pinpoint the issue, enabling targeted corrections~\cite{hardt2016equality}.

Another detection method is \textit{Dataset Visualization Techniques}, such as t-SNE (t-Distributed Stochastic Neighbor Embedding) and PCA (Principal Component Analysis), which allow researchers to examine bias in high-dimensional healthcare data in more intuitive ways. These tools uncover patterns, such as clusters or representation gaps, that may signal clinical representation bias. For example, a t-SNE plot could reveal that certain skin tones are underrepresented in a dermatological dataset, prompting actions to balance the dataset~\cite{vandermaaten2008tsne}. Another approach is the use of \textit{Causal Graphs}, which focus on identifying and understanding the relationships between variables in healthcare datasets. Causal Graphs are graphical representations of causal relationships among variables, allowing researchers to distinguish true causation from correlations and uncover confounding variables that may introduce bias. For instance, \cite{grari2022fairness} proposed a framework leveraging Causal Variational Autoencoders (CVAEs) to indirectly reconstruct sensitive information, even when such attributes are unavailable due to privacy constraints. By identifying clinical biases at the data level, this method proves particularly valuable in addressing challenges within complex, real-world medical datasets.

\subsubsection{In-Processing Stage}

The in-processing stage focuses on detecting biases introduced or amplified during model training. This stage prioritizes fairness by incorporating bias identification strategies directly into the training process. One effective method is \textit{Adversarial Learning}, which employs adversarial techniques to identify bias in models during training. A primary model predicts clinical outcomes (e.g., disease risk ), while an adversary attempts to infer sensitive attributes such as patient demographics (e.g., race, gender) based on the model’s predictions. The adversary’s success in predicting these attributes reveals the extent of algorithmic bias embedded in the model. By analyzing this feedback, healthcare developers can identify where and how protected attributes influence clinical predictions, enabling precise bias detection and promoting equitable care~\cite{zhang2018biasadversarial}.

Another approach involves \textit{Explanation Methods}, such as LIME (Local Interpretable Model-Agnostic Explanations), which provide insights into how clinical features influence predictions in healthcare models. These methods can uncover both explicit and implicit biases in the model, such as disproportionately relying on sensitive attributes like race, gender, or socioeconomic status or unintentionally reinforcing biased patterns that disadvantage certain patient groups. By iteratively applying these explanations during training, healthcare practitioners can identify and correct biases, improving fairness and transparency in patient care~\cite{ribeiro2016should}. A further technique for bias detection is \textit{Causal Modeling}, which identifies algorithmic bias by analyzing how sensitive attributes, such as gender, ethnicity, or socioeconomic status, influence predictive outcomes in healthcare applications. Causal Modeling involves constructing mathematical or computational models that map out causal relationships between variables, enabling researchers to trace pathways where biases emerge during the training of models used in medical decision-making.  For instance, researchers utilized causal modeling with the Adult dataset from the UC Irvine Machine Learning Repository to detect gender bias in binary classification outcomes~\cite{hui2023detecting}. By treating the prediction model as a black box and constructing a corresponding causal model, they identified how gender impacted the model's decisions. This process uncovers both explicit and implicit biases in the model’s decision-making framework, providing actionable insights for detecting inequities. By identifying biases at the model level, causal modeling promotes the development of fair and robust AI systems in health care.

\subsubsection{Post-Processing Stage}

The post-processing stage evaluates and mitigates any biases in the model’s outputs after training is complete. This stage ensures equitable treatment in the model’s final decisions. One effective approach for bias detection is \textit{Counterfactual Analysis}, which assesses whether a model’s decisions remain consistent even if sensitive attributes, such as race or gender, are changed. By using causal inference, this method determines whether changing a sensitive attribute would alter the model’s outcome. If the outcome remains the same, the decision is considered fair, making this method effective for identifying and correcting both explicit and implicit biases in the model’s outputs~\cite{kusner2018counterfactual} For instance, in a healthcare model that predicts eligibility for advanced treatments, if changing a sensitive attribute like race results in different treatment outcomes, it highlights bias in the model's decision-making process.

Another tool is \textit{Model Cards}, which are standardized documents that summarize a model’s performance across various clinical conditions and demographic groups. They include details about the model’s intended use in healthcare settings, limitations, and evaluation results, including the detection and analysis of evaluation bias. By promoting transparency, Model Cards help healthcare stakeholders make informed decisions about deployment and encourage responsible use~\cite{mitchell2019modelcards}. Another detection approach, \textit{Reject Option} involves identifying cases near the decision boundary where classification decisions are uncertain and potentially influenced by bias. This approach helps detect patterns where sensitive demographic groups might be disproportionately affected by misclassifications. In healthcare settings, where errors in diagnosis or treatment can have significant consequences, the reject option highlights instances that require further inspection to ensure equitable outcomes. By focusing on borderline cases, this method provides valuable insights into potential biases in model predictions, aiding in the identification of inequities within healthcare systems~\cite{kamiran2018rejectoption}.

\bigskip

\noindent This framework provides a clear and actionable approach to tackling biases in AI systems. Detecting and identifying biases at each stage of the ML pipeline is essential for creating fair, reliable, and equitable AI-driven healthcare solutions.

\subsection{Mitigation Strategies}

Mitigating biases in AI systems within healthcare is crucial for ensuring equitable treatment across all patient demographics. This section outlines key strategies to address and reduce biases, including the use of diverse and representative data and fairness-aware approaches, thereby enhancing the fairness and accuracy of AI models. 

\subsubsection{Diverse and Representative Data}
Mitigating biases in AI systems within healthcare is crucial for ensuring equitable treatment across all patient demographics. A fundamental strategy involves using diverse and representative datasets in the training phase of these systems. Research indicates that AI models can only be as unbiased as the data they are trained on~\cite{obermeyer2016predicting}. Thus, the inclusion of comprehensive data from a wide array of patient groups, particularly those historically underrepresented in medical research, is essential~\cite{chen2019can}.

The significance of representative data extends beyond just including diverse groups; it also involves the depth and quality of the data collected from these groups. Data must capture a breadth of variables that influence health outcomes, such as socioeconomic factors, environmental conditions, and genetic differences~\cite{rajkomar2019machine}. This approach helps in creating models that are better tuned to the nuances of various patient needs and conditions~\cite{adamson2018machine}.

Additionally, involving stakeholders from diverse backgrounds in the data collection and AI development process can enhance the relevancy and sensitivity of the datasets~\cite{vyas2020hidden}. This inclusion helps in identifying and addressing potential blind spots in AI training datasets, which, if overlooked, could perpetuate biases and inequalities in healthcare delivery~\cite{gebru2021datasheets}. Employing these strategies not only improves the fairness and effectiveness of AI tools but also builds trust in these technologies among all user groups, thereby fostering a more inclusive healthcare ecosystem~\cite{char2018implementing}.

\subsubsection{Fairness-aware approaches}
Incorporating various fairness-aware approaches is crucial for developing AI systems that are equitable and unbiased. These approaches can be categorized into three main strategies: pre-processing, in-processing, and post-processing. Each strategy targets different stages of the machine learning pipeline to address potential biases and ensure fair treatment across diverse demographic groups.

\paragraph{Pre-Processing:} In healthcare, pre-processing involves techniques to adjust the input data for AI models to ensure it accurately represents diverse patient demographics before model training begins. This process aims to eliminate any inherent biases that may skew AI predictions and outcomes, thus fostering equity in healthcare treatments and diagnostics~\cite{kamiran2012data}. By refining the dataset upfront, pre-processing helps in building AI systems that perform fairly across all patient groups~\cite{chen2019can}.
\begin{itemize}
    \item \textit {Re-Sampling:} Re-sampling is a common pre-processing method that manipulates data samples to create a more balanced dataset. This can be done through either over-sampling minority classes or under-sampling majority classes~\cite{lum2016statistical}. An example of this in healthcare is the use of the Synthetic Minority Oversampling Technique (SMOTE)~\cite{chawla2002smote} in medical datasets where certain conditions or outcomes are rare. Chawla applied SMOTE to a dataset of imbalanced classes to improve the prediction of minority-class outcomes in medical diagnostics, resulting in more reliable predictive models for underrepresented conditions. This technique helps to ensure that the AI does not become biased towards the more frequent classes.
    \item \textit{Re-Weighting:} To reduce bias, instances in the training data are reweighted. We assign a weight to each instance in a dataset based on its representation, thereby increasing the influence of underrepresented groups in the model training process. This technique was demonstrated by Calmon in~\cite{calmon2017optimized}, who developed a data pre-processing method that modifies features and outcomes to improve fairness. In healthcare, this could mean adjusting weights so that data from minority ethnic groups has a greater influence on the model, helping to prevent the AI from developing biases that might affect diagnosis or treatment recommendations~\cite{kamiran2012data}.
\end{itemize}

\paragraph{In-Processing:} In-processing methods involve integrating fairness directly into the learning algorithm itself. This technique adjusts the algorithm during the training phase to minimize bias. Zemel in his study introduced a method where they developed a representation for data that is invariant to protected attributes like race or gender while maintaining the informative characteristics necessary for prediction~\cite{zemel2013learning}. In healthcare, this can be crucial for ensuring that diagnostic or treatment recommendations are not adversely skewed by underlying biases in the training data~\cite{deldjoo2021survey}.

\begin{itemize}
   \item \textit{Adversarial Debiasing:} Adversarial debiasing involves the simultaneous training of a predictor and an adversary. The predictor learns to make accurate predictions, and the adversary learns to determine whether the predictions are biased toward certain groups. Deldjoo et al.~\cite{deldjoo2021survey} explored this in the context of healthcare, aiming to create clinical decision support systems that are unbiased towards patients' demographic attributes. This technique helps ensure that the AI's treatment recommendations do not reflect discriminatory patterns that might exist in the training data~\cite{zafar2017fairness}.
   \item \textit{Constraint-Based Optimization:} Constraint-based optimization involves modifying the learning algorithm to satisfy fairness constraints while minimizing prediction error. You can implement this method by adding fairness constraints to the objective function that the algorithm is optimizing. Elzayn et al.~\cite{elzayn2019fair} applied this method to healthcare datasets, ensuring that treatment recommendations do not disproportionately favor one group over another by enforcing demographic parity or equality of opportunity as constraints during model training~\cite{donini2018empirical}.
\end{itemize}

\paragraph{Post-Processing:} After training a model, post-processing techniques are applied to modify the outputs of AI systems, ensuring fair treatment across different demographic groups. These adjustments address any residual biases to promote equitable outcomes. For example, Hardt et al.~\cite{hardt2016equality} applied different decision thresholds for various groups to equalize treatment effects. This method effectively adjusted risk scores in a healthcare dataset, ensuring that treatment recommendations were equitable~\cite{pleiss2017fairness}.
\begin{itemize}
    \item \textit{Threshold Adjustment:} Threshold adjustment in AI healthcare involves setting an optimal cutoff point for predictive models to classify outcomes effectively, such as predicting disease presence or patient risk levels. A significant example of this is in cardiovascular disease prediction. In a study by Siontis~\cite{siontis2018outcomes}, the researchers examined the application of different thresholds for predicting cardiovascular events using a logistic regression model. They aimed to optimize the sensitivity and specificity of predictions by adjusting the threshold to better serve clinical decision-making. This adjustment proved crucial in identifying higher-risk patients who might benefit from preventive treatments. Another notable case is in breast cancer screening, where Fenton~\cite{fenton2013short} applied threshold adjustment to mammography interpretation models. By altering the decision threshold, they were able to reduce false positives without substantially missing cases of cancer, thereby balancing the need for early detection with the risk of overdiagnosis and unnecessary anxiety for patients.
    \item \textit{Output Recalibration:} Output recalibration in AI healthcare is critical for adapting predictive models to local contexts and ensuring that the predicted probabilities match actual clinical outcomes in a new patient population. A notable case involves the recalibration of a sepsis prediction model at Johns Hopkins Hospital, as detailed by Henry et al.~\cite{henry2015hospital}. The model was originally developed with a dataset from one patient demographic, but recalibration using local demographic data was required to maintain accuracy across diverse populations within the hospital. This recalibration helped align the model's predictions with the actual rates of sepsis observed, improving clinical decision-making and patient care. Another example is from the University of California, San Francisco (UCSF), where Rajkomar~\cite{rajkomar2019machine} recalibrated Google’s deep learning model used for predicting medical outcomes. The model, initially trained on a national dataset, was recalibrated with UCSF-specific patient data to ensure its predictions accurately reflected the local patient population’s risk factors and outcomes. This was particularly important for diseases with variable manifestations across different demographics.
\end{itemize}

\section{Research gaps and Future directions}
\label{sec:sec_4}

This section delves into the research gaps in implementing fair AI in healthcare and future directions to enhance its efficacy. Ensuring fairness in AI systems within healthcare is a complex and multifaceted challenge that requires addressing various deficiencies and promoting interdisciplinary collaboration.

\subsection{Ethical Considerations}

The integration of AI in healthcare raises significant ethical challenges, including privacy concerns, algorithmic bias and accountability. One critical gap lies in ensuring patient data privacy while enabling AI models to be trained on diverse datasets. Ensuring the confidentiality and security of this data against breaches is paramount~\cite{mittelstadt2019principles}. Current approaches, such as anonymization, may not adequately prevent re-identification risks, particularly when combined with external data sources~\cite{lubarsky2017ReIdentification}. Future research should explore privacy-preserving techniques, such as federated learning and differential privacy, to enable secure training of AI systems without compromising data confidentiality~\cite{ling2024FedFDP}. Algorithmic bias is another major concern. AI models trained on non-representative data can perpetuate systemic inequalities, leading to unfair treatment outcomes for underrepresented groups~\cite{rajkomar2019machine}. Solutions like adversarial debiasing and fairness-aware algorithms should be further developed and rigorously tested in healthcare settings~\cite{poulain2023FairnessFL}. Furthermore, the deployment of AI in healthcare must take into account the potential displacement of healthcare professionals, raising concerns about job security and the loss of valuable human expertise and empathy in patient care~\cite{davenport2019potential}. Lastly, accountability for AI-driven decisions in healthcare is a critical ethical issue. It is critical to define who is responsible—the developers, the users, or the AI itself—when an AI system's decision results in patient harm~\cite{luxton2014artificial}.

\subsection{Legal Considerations}

Legal considerations in AI healthcare systems encompass intellectual property rights, liability for errors, regulatory compliance, and policy gaps that hinder bias mitigation. A notable gap exists in the legal frameworks governing intellectual property rights as there is an ongoing debate over whether AI-generated medical inventions can be patented and who owns the rights to AI-generated data and algorithms~\cite{price2019privacy}. Future research should focus on developing frameworks that equitably address the rights and interests of developers, institutions, and patients. Furthermore, liability issues arise when AI systems make erroneous decisions that could harm patients. Determining whether the healthcare provider, the software developer, or another party is liable requires intricate legal analysis and possibly new legal frameworks~\cite{hall2018medical}. Solutions such as liability insurance for AI developers and a staggered approach to liability, from strict liability for high-risk applications to simple fault-based liability for consumers, could address this gap~\cite{hacker2022ai}. Policy gaps further hinder bias mitigation, particularly when demographic information is incomplete or unavailable. Traditional fairness approaches in AI healthcare systems often operate under the complete demographic assumption, defining a limited set of groups based on gender, race, or other protected attributes, and enforcing similar outcome statistics across these predefined groups~\cite{wang2025towards}. However, in practice, this assumption frequently fails due to privacy, legal, and regulatory constraints that limit access to comprehensive demographic information~\cite{wang2025towards_fairness}. These limitations create significant challenges for implementing fairness in real-world AI healthcare applications, necessitating novel technical approaches that can achieve fairness without complete demographic data~\cite{wang2025fairgnn}. Future directions also include promoting policy changes that enforce transparency and fairness in data collection and algorithmic processing in healthcare applications~\cite{char2018implementing}. Governments and institutions should implement policies mandating fairness audits and bias detection tools to ensure equitable outcomes. AI applications in healthcare must adhere to existing healthcare regulations, such as the Health Insurance Portability and Accountability Act (HIPAA) in the U.S., which governs the privacy and security of patient data, and the General Data Protection Regulation (GDPR) in the EU, which sets stringent guidelines for data protection and privacy~\cite{terry2017regulatory}. As AI technologies evolve, there may also be a need for specific regulations that address AI’s unique aspects in healthcare to ensure that these innovations are safely and effectively integrated into medical practice~\cite{gerke2020ethical}. Lastly, issues of informed consent are magnified with AI, as patients must understand the role of AI in their care, including how their data will be used and the implications of AI-driven decisions~\cite{luxton2014artificial}.

\nocite{*}

\subsection{Transparency and Interpretability}

The lack of transparency and interpretability in AI models remains a major obstacle to fairness in healthcare. Many advanced models, such as deep neural networks, function as "black boxes," making their decision-making processes hard to understand and verify. This lack of clarity can lead to mistrust among healthcare providers and patients, as well as unidentified biases in outcomes~\cite{holzinger2017we}. For instance, an AI diagnostic tool might predict a higher risk of disease for one patient over another with nearly identical medical histories, without providing any rationale for the differing assessments. To address this, future research should focus on developing “glass-box” approaches that are inherently interpretable, such as decision trees or linear models, fostering greater trust among healthcare providers and patients~\cite{holzinger2017we}. Future research should focus on adopting model-agnostic explanation frameworks like LIME~\cite{ribeiro2016should} or SHAP~\cite{ lundberg2017unified}, which provide insights into the decision-making processes of any AI model, regardless of its underlying architecture. Developing these tools will enable clinicians to better understand and oversee AI-driven decisions, potentially leading to more equitable healthcare outcomes. Additionally, implementing model documentation standards like Model Cards can provide essential transparency by detailing the capabilities and limitations of AI models to all stakeholders~\cite{mitchell2019modelcards}.

\subsection{Diversity and Representation}

The lack of diversity in training datasets is a significant research gap that impacts AI fairness in healthcare. Models trained on data that underrepresents certain demographic groups risk perpetuating biases, thereby worsening healthcare inequalities~\cite{rajkomar2019machine}.  For example, a skin cancer detection AI trained predominantly on lighter skin tones may fail to accurately identify conditions on darker skin, highlighting the impact of underrepresentation in training data on healthcare outcomes. To mitigate this issue, techniques for synthetically augmenting underrepresented classes in datasets or incentivizing the collection of more comprehensive and inclusive data must be developed~\cite{chen2018my}. Additionally, implementing algorithmic fairness approaches during model training, such as fairness constraints or adversarial debiasing, can help mitigate biases introduced by skewed datasets~\cite{mehrabi2021survey}.  Studies suggest that geospatial AI models, when applied to geographic information systems (GIS), can uncover regions with inadequate healthcare services, thus guiding interventions to those most in need~\cite{veinot2018good}. However, predictive analytics must be carefully designed to ensure they do not perpetuate biases present in underlying data~\cite{obermeyer2019dissecting}. Integrating feedback mechanisms from healthcare providers into AI systems offers a pathway to refine these models, ensuring they adapt to serve diverse populations effectively. The future direction in bridging this gap involves developing AI technologies that are not only predictive but also prescriptive, providing actionable insights to improve universal healthcare accessibility~\cite{rajkomar2019machine}.

\subsection{Interdisciplinary Approaches}

The integration of AI in healthcare requires interdisciplinary approaches that combine insights from medical ethics, data science, social sciences, and clinical practice to comprehensively address fairness~\cite{goodman2017european}. However, current AI applications often lack this multidimensional perspective, resulting in solutions that are technically adequate but socially or ethically inappropriate~\cite{char2018implementing}.  For example, an AI diagnostic tool developed with only technical expertise might fail to account for socio-economic factors, leading to misdiagnoses in underserved communities that lack access to quality healthcare infrastructure. Future research should focus on frameworks that incorporate not only technical accuracy but also ethical, legal, and social implications from the outset, ensuring that AI systems are developed with a holistic view of fairness~\cite{veinot2018good}. These collaborative frameworks can help prevent the exacerbation of existing disparities and promote a more equitable distribution of healthcare resources~\cite{chen2019can}.

\subsection{Longitudinal Effects}

The longitudinal effects of AI systems on healthcare fairness remain underexplored, with existing studies often focusing on short-term outcomes and immediate biases in algorithmic decision-making~\cite{ghassemi2019practical}. Understanding how these technologies affect health disparities over time is crucial, as initial biases can amplify if not properly addressed~\cite{rajkomar2019machine}. For example, an AI system predicting patient readmissions might keep giving more resources to well-funded hospitals because they had better outcomes in the past, making the gap worse for less-supported hospitals over time. Future research should involve continuous monitoring of AI systems post-deployment to assess their impact on various demographic groups across different time intervals~\cite{beam2018big}. This approach will help identify evolving biases and enable timely modifications to algorithms, thereby ensuring more equitable healthcare outcomes~\cite{obermeyer2019dissecting}. Additionally, incorporating feedback loops that allow healthcare providers to report disparities can further refine AI applications in real-world settings.

\section{Conclusion}
\label{sec:sec_7}

In conclusion, AI in healthcare holds transformative potential for improving diagnosis, treatment, and patient management. However, its integration also brings significant ethical, legal, and operational challenges, especially regarding fairness and equity. To balance the benefits of AI with these challenges, it is essential to address biases, ensure ethical practices, and foster interdisciplinary collaboration. Addressing biases involves implementing more inclusive data collection practices to ensure diverse representation and developing bias-aware AI models to mitigate disparities in healthcare delivery. Ensuring ethical practices requires establishing robust regulatory frameworks to oversee AI implementation, guaranteeing compliance with ethical standards and legal requirements. Additionally, promoting transparency in AI decision-making processes is essential to building trust among healthcare providers and patients. To achieve these goals, fostering interdisciplinary collaboration is vital. Encouraging ongoing cooperation between technologists, clinicians, legal experts, and ethicists will help address the multifaceted challenges of AI in healthcare. Interdisciplinary research should focus on developing innovative and inclusive AI applications that effectively serve diverse populations without perpetuating existing disparities. Future research should aim to refine AI tools to ensure their effectiveness across diverse populations, continuously monitoring and adjusting to evolving biases.
Investigating the long-term impacts of AI systems on healthcare equity is crucial for making necessary adjustments over time. By addressing these critical areas, we can harness the capabilities of AI to create a more equitable and effective healthcare system. As we advance, it is imperative to remain vigilant about the ethical implications of AI, ensuring that these technologies benefit all segments of society equally. This requires a committed effort to integrate fairness, transparency, and inclusivity into every stage of AI development and deployment in healthcare.

\section*{Acknowledgements}

This work was supported in part by the National Science Foundation (NSF) (Grant No. 2404039) and the National Institutes of Health (NIH) (Grant No. R01MD019814). 


%
%
%

\end{document}